# Very strict selectional restrictions: a comparison between Portuguese and French

Éric Laporte, Christian Leclère

*IGM, Université de Marne-la-Vallée - 5, bd. Descartes*
*77454 Marne-la-Vallée CEDEX 2 -France*
eric.laporte@univ-mlv.fr - christian.leclere@univ-mlv.fr

Maria Carmelita Dias

*Pontifícia Universidade Católica do Rio de Janeiro*
*Rua Marquês de S. Vicente, 225 - Rio de Janeiro - 22453-900 - Brasil*
mcdias@let.puc-rio.br

This work has been partially supported by the CNRS.

## 1. Introduction

We discuss the characteristics and behaviour of two parallel classes of verbs in two Romance languages, French and Portuguese. Examples of these verbs are Port. *abater [gado]* and Fr. *abattre [bétail]*, both meaning 'slaughter [cattle]'. Such collocations are intermediate cases between verbal idioms and largely free verb phrases. Precise knowledge of these verbs would aid recognition of verb senses in texts and therefore be useful for natural language processing. The objective of this study is to compare the importance of these classes of verbs within the respective lexicon of both languages, and in particular to investigate corresponding pairs such as *abater [gado]/abattre [bétail]*.

## 2. Related work

The distributional restrictions attached to word combinations show a continuum ranging from relatively free combinations between meaningful words (e.g. *mount a hill*) to the unique combinations observed between the elements of idioms (e.g. *mount the guard*). A distributionally frozen idiom such as *mount the guard* can be seen as an extreme case of selectional restriction in which only *the guard* can fill one of the slots of the verb (Guillet, 1986). In the intermediate case, the set of nouns that can fill a slot is reduced to a semantically homogeneous set of few nouns. For example, the verb *mount* in *mount a show* has a precise meaning observed only with *play, opera* and a few other nouns denoting spectacles. A list of phrases corresponding to this situation, called 32R3, had actually been produced for French (Boons *et al.*, 1976), and was extended later by the LADL[1] to a list of 1003 verbs[2].

When a verb selects a small set of nouns as its possible complement, this set can sometimes be characterized as the set of the hyponyms of a noun. For example, all the complements of Port. *abater [avião, helicóptero]* 'bring down [plane, helicopter]' are probably *aeronave* 'aircraft' and its hyponyms. Thus, our research is also indirectly connected with ontologies (Gruber, 1993) and the semantic classification of nouns (Gross, Clas, 1997).

## 3. Comparison between French and Portuguese

We borrowed the definition of French class 32R3 from Leclère (2002) and transferred it to Brazilian Portuguese[3]. Verbs in this class share some important features: e.g. they take only one essential complement and do not admit a sentential object or subject. These criteria define the classes as highly residual ones, i.e. they select verbs with few other interesting properties[4].

We made a cross-lingual comparison between samples of the French and Portuguese classes. Our starting point was the list of French verbs (Boons *et al.*, 1976) beginning with *a*, in its updated version, and their translation into Portuguese. We then looked up all the verbs starting with *a* in an important Portuguese dictionary (Houaiss, 2001) and satisfying the definition, and translated the Portuguese verbs into French. This process produced a list of 74 Portuguese candidates and 75 French candidates to be included in the class. These lists include the entries in *a* that are members of the class in one of the languages, and their translation into the other, if they exist[5]. However, collocations and other word combinations are language-dependent phenomena: even if a member of the class in one language has a translation into the other, this translation is not necessarily a member of the class in the other language. For example, Fr. *aplanir [difficulté]* can be translated into Port. *amainar [dificuldade]*, but Port. *amainar* can be used with much more various complements than Fr. *aplanir* (with this abstract meaning). Other verbs are only translated by syntactically different constructions[6]. Thus, among a total of 74 pairs (French and Portuguese verbs of the sample and their translation), 47 % are pairs of members of this class in both languages (see table).

---

[1] Laboratoire d'automatique documentaire et linguistique, Université Paris 7, 1968-2000.
[2] This list is available on the web at http://infolingu.univ-mlv.fr/english: follow Linguistic Data, Lexicon-Grammar, View.
[3] When a verb has several senses, we considered each of them as a separate lexical entry, e.g. Port. *abater [avião]* 'bring down [plane]' was considered distinct from *abater [gado]* 'slaughter [cattle]'.
[4] However, the phenomenon of very strict selection also occurs with other verbs, e.g. intransitive or ditransitive verbs, and is worth a systematic study.
[5] Fr. *aplatir [balle]*, a technical term of rugby, has no translation in Brazilian Portuguese.

Table – Verbs with very strict selectional restrictions in French and Portuguese (excerpt)

| FRENCH | | PORTUGUESE | |
|---|---|---|---|
| VERB | COMPLEMENT | VERB | COMPLEMENT |
| abattre | avion... | abater | avião... |
| abattre | bétail... | abater | gado... |
| abjurer | foi... | abjurar | fé... |
| abolir | loi... | abolir | lei... |
| abroger | loi... | ab-rogar | lei... |
| boucler | ceinture... | afivelar | cinto... |

We found that 62 % of the 74 Portuguese verbs, and 85 % of the 75 French verbs could be included into the class. This difference tends to indicate that though it is present in both languages, the phenomenon could affect more entries in French than in Brazilian Portuguese.

## 4. The sets of nouns selected

The verbs in class 32R3 and in its Portuguese counterpart select small sets of nouns. These sets of nouns are by definition semantically homogeneous. For example, the complements of Port. *abater [avião]* 'bring down [plane]' are *aeronave, avião, helicóptero* and others. Some of these sets of nouns are even sets of synonyms: for example, Port. *amortecer [impacto]* takes such complements as *choque, colisão, impacto...* How can these sets of nouns be represented in a formal grammar? The conclusions of Guillet (1986) about French can be extended to Portuguese. Some of the sets of nouns in question can be specified through the choice of canonical representatives, e.g. *aeronave* for the complements of *abater [avião]*, i.e. these sets consist of the canonical representative and all its hyponyms. For example, the set of hyponyms of *aeronave* is the set of nouns $N$ for which the sentence *N é um aeronave* 'N is an aircraft' is true for the common sense[7]. However, in some cases, the only suitable canonical representative would be a whole phrase including elements of definition. For example, the complements of Fr. *abattre [bétail]* 'slaughter [cattle]' denote domestic animals bred for certain purposes, but no French noun means this: *bétail* 'cattle', which excludes poultry, is two narrow, and *animal domestique* 'domestic animal', which includes pets, is too general. In other cases, it is difficult to find even a periphrasis that adequately defines the set of nouns. This is the case of Fr. *applaudir [spectacle, discours]* and Port. *aplaudir [espetáculo, discurso]* 'applause [show, speech]', though the meaning of the verb is the same for a show and for a speech.

---

[6] Examples: without complement (Port. *abortar [feto]*/Fr. *avorter*), with a prepositional object (Fr. *amender [loi]*/Port. *fazer uma emenda em [lei]*), or with two verbs (Port. *adernar [embarcação]*/Fr. *faire gîter [embarcation]*).

## 5. Conclusion

The class of verbs we discussed are quite frequent in Portuguese and French, and probably also in other languages. Therefore, their study is useful for a number of computer applications both monolingual and bilingual. We showed that the behaviour of these combinations in Portuguese and in French are quite similar, but that the combinations themselves are mirrored by equivalent ones in roughly half the cases. The sets of nouns selected by this type of verbs are interesting subjects of study themselves, but their representation by canonical representative is bound to be only an approximation in a large number of cases.

---

[7] This set certainly includes *avião* and *helicóptero*, but relevant questions are: does the set of hyponyms include all the complements of *abater [avião]*? do all the complements of *abater [avião]* belong to the set of hyponyms? If the answer to both questions is yes, *aeronave* becomes a good choice to define this lexical entry of *abater*.